\documentclass{article}

\usepackage[preprint]{neurips_2023}

\usepackage[utf8]{inputenc} % allow utf-8 input
\usepackage[T1]{fontenc}    % use 8-bit T1 fonts
\usepackage{hyperref}       % hyperlinks
\usepackage{url}            % simple URL typesetting
\usepackage{booktabs}       % professional-quality tables
\usepackage{amsfonts}       % blackboard math symbols
\usepackage{nicefrac}       % compact symbols for 1/2, etc.
\usepackage{microtype}      % microtypography
\usepackage{xcolor}         % colors
\usepackage{graphicx}
\usepackage{float}

\title{AI4GCC - Track 3: Consumption and the Challenges of Multi-Agent RL}

\author{%
  Marco Jiralerspong\\ %\thanks{} \\
  Mila, University of Montreal\\
\texttt{marco.jiralerspong@mila.quebec} \\
  \And
  Gauthier Gidel \\
  Mila, University of Montreal \\
  Canada CIFAR AI Chair \\
  \texttt{gidelgau@mila.quebec} \\
}

\begin{document}

\maketitle

\begin{abstract}
    The AI4GCC competition presents a bold step forward in the direction of integrating machine learning with traditional economic policy analysis. Below, we highlight two potential areas for improvement that could enhance the competition's ability to identify and evaluate proposed negotiation protocols. Firstly, we suggest the inclusion of an additional index that accounts for consumption/utility as part of the evaluation criteria. Secondly, we recommend further investigation into the learning dynamics of agents in the simulator and the game theoretic properties of outcomes from proposed negotiation protocols. We hope that these suggestions can be of use for future iterations of the competition/simulation.
\end{abstract}

% \section{Introduction}
% The AI4GCC competition presents a bold step forward in the direction of integrating machine learning with traditional economic policy analysis. Below, we highlight two potential areas for improvement that could enhance the competition's ability to identify and evaluate proposed negotiation protocols. Firstly, we suggest the inclusion of an additional index that accounts for consumption/utility as part of the evaluation criteria. Secondly, we recommend further investigation into the learning dynamics of agents in the simulator and the game theoretic properties of outcomes from proposed negotiation protocols. We hope that these suggestions can be of use for future iterations of the competition/simulation.
\section{Evaluating Consumption}

The first area of improvement is the disconnect between the indices used to evaluate the solution and the reward being maximized by the reinforcement learning (RL) agents. Submissions are evaluated on two indices: the \textbf{economic index} (based on gross output, designed to evaluate economic productivity) and the \textbf{climate index} (based on temperature rise, designed to evaluate environmental impact) ~\citep{zhang2022ai,Zhang_RICE-N_2022}. Ideal solutions will obtain high values for both, maintaining relatively high economic productivity while having minimal impact on the environment. Ultimately however, there is a tradeoff between the two: increasing economic productivity tends to worsen environmental impact and vice versa.

\begin{figure}[h]
  \centering
  \includegraphics[width=0.5\textwidth]{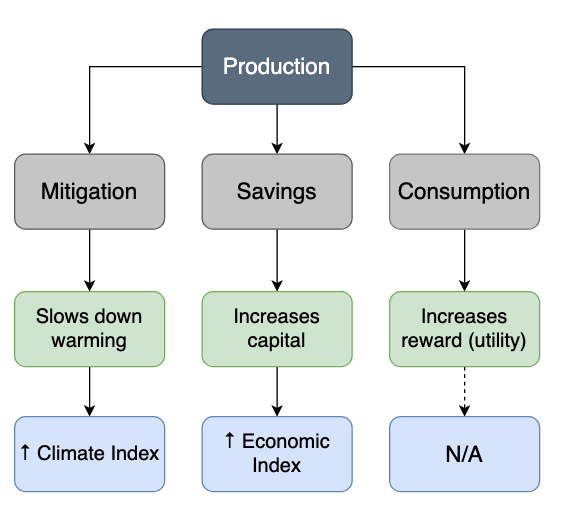}
  \caption{Allocation decision at each step. Each country produces a certain amount of output that is either used for mitigation, savings or consumption.}
  \label{fig:production_consumption_savings}
\end{figure}

However, the tradeoff being faced by the RL agents is not simply between protecting the environment and producing a large amount of goods but along a third dimension as well: \textbf{consumption}. Given their output for a period, agents can either invest it (growing their capital and allowing them to produce more in the future), spend it on mitigation (allowing them to reduce their impact on the climate) \textit{or} consume it (increasing their utility). Of the three, consumption (of domestic and foreign goods) is the only element that directly affects agent reward. Agents are only indirectly interested in improving their economic output or their environmental impact if it allows them to consume more.

Yet, consumption is not being evaluated by the two indices. As such, it is possible to improve both the economic and climate indices \textit{at no cost} (or at least no cost taken into account by the current evaluation scheme) by simply reducing consumption. In fact, throughout training, as agents are just rewarded for consumption, they will often learn to increase their consumption at the expense of economic output and mitigation. As a result, after training, the simulation outcomes are often worse for these two indices than before training while agent reward is much higher. The current evaluation scheme thus tends to reward submissions where agents have not been trained fully rather than submissions where agents have appropriately learned to negotiate (but also increase their consumption), as the competition intended.

As examples, we include in Table 1 the key metrics for the no negotiation baseline (before and after 10k episodes of training) as well as an artificially extreme example where agents are forced to invest/mitigate the maximum amount (through an action mask), leaving no room for consumption.\footnote{There is still some consumption left as we believe there is a minor bug in the simulation with the number of discrete action levels (should divide by the number of discrete action levels - 1).}

\begin{table}[h]
    \label{tab:metric_examples}
    \centering
    \begin{tabular}{ l c c c} 
        \toprule
        & \textbf{No negotiation (before)} & \textbf{No negotiation (after)} & \textbf{No consumption} \\ 
        \midrule
        \textbf{Capital} & 33662  & 13441 & \textbf{86926}  \\
        \addlinespace
        \textbf{Gross Output} & 6704 & 5364 & \textbf{8665} \\
        \addlinespace
        \textbf{Temperature rise} & 4.51 & 4.55 & \textbf{2.37}\\
        \addlinespace
        \textbf{Economic index} & 0.67 & 0.52 & \textbf{0.89} \\
        \addlinespace
        \textbf{Environmental index} & 0.43  & 0.43 & \textbf{0.81}\\
        \addlinespace
        \textbf{Episode reward} & 104.62 & \textbf{164.98} & 15.17\\
        \bottomrule
    \end{tabular}
    \vspace{1em}
    \caption{Key metrics for the no negotiation baseline before training (left), the no negotiation baseline after training (middle) and the artificial example with no consumption (right). Crucially, the submission obtains better metrics before instead of after training for all metrics except episode reward (which increases as agents learn to tradeoff optimally between savings and consumption, at the expense of the other two indices). These values are dwarfed by the no consumption case where agents obtain very high values for both indices but terrible rewards. If episode reward is not evaluated, the no consumption submission looks by far to be the best even though it completely ignores utility.}
\end{table}

\vspace{-1em}

Agent reward/utility is a crucial aspect of this environment and is necessary for the interplay between trading/tariffs and mitigation. The benefits of trade are only reflected through the utility of the agents (which is higher for some amount of trade due to the Armington assumption). Without it, agents cannot reward/punish each other for mitigating/not mitigating through trade which is the fundamental dynamic behind these negotiation protocols and what agents are supposed to be learning.

To remedy this issue, we suggest evaluating submissions along this third axis with a \textbf{utility index} (or replacing the economic index with a utility index), rewarding submissions where agents are able to achieve high utility while also mitigating.

\section{Multi-Agent RL Challenges}
Secondly, a more high-level concern is the lack of research on the performance of multi-agent RL (MARL) for environments with free-riding concerns. Just single agent deep-RL can, by itself, be notoriously hard to train with a strong dependence on initialization, randomness and hyperparameter selection. Adding multiple heterogeneous agents that need to manage the interplay between cooperation/competition complicates an already difficult problem. As such, it may be unwise to place excessive trust in simple MARL (based on self-play) to converge to solutions that are optimal and stable given a negotiation protocol.

\subsection{Convergence to Optimal Nash Equilibria}
% Show that even in a simple case, RL algorithms might not converge to a Nash equilibrium
The first potential issue with MARL in free-rider games is examining if agents converge to a Nash equilibrium. Specifically, under the current setup, there is no evaluation of whether the outcome achieved is vulnerable to deviation from agents. If the outcome is not a Nash, it is unlikely the solution will be attained in a real world setting as rational agents will deviate to improve their payoff.

One way to evaluate this is to fix the trained RL agents and loop over countries in the simulation. For each, a separate agent could be trained independently against the fixed agents. If it manages to substantially improve its reward, the RL agents did not converge to a Nash equilibria. If not, the solution is likely resistant to individual defection. Ideally, the solution should also be checked for a coalition resistance (i.e. could a coalition benefit by defecting) but this is a combinatorially large problem. Nonetheless, some estimate could be obtained by, for example, sampling random subsets of agents and training them against the fixed agents.

Even if countries converge to a Nash equilibrium, it is possible that many equilibria exist (as in the iterated prisoner's dilemma case). Getting RL agents to converge to a cooperative equilibrium may be challenging with uncooperative equilibria being reached instead (e.g. converging to always defect when the other equilibria mentioned above have higher payoffs). As such, potentially viable negotiation protocols might be skipped over as RL agents fail to learn cooperative equilibria they induce.

We demonstrate these failure cases in a very simple environment that exhibits similar free-riding dynamics: the infinite horizon iterated prisoner's dilemma (see \ref{Appendix} for details). Specifically, we replicate some of the findings of ~\citep{foerster2017learning, lerer2017maintaining} using an A2C agent setup as in ~\citep{Zhang_RICE-N_2022}. As state, the agent only has access to the outcome of the last timestep and must choose between cooperation and defection. Even in this very simple environment, both issues arise. As such, it would be prudent to investigate these further in the significantly more complex AI4GCC environment.

\begin{figure}[h]
  \centering
  \includegraphics[width=1\textwidth]{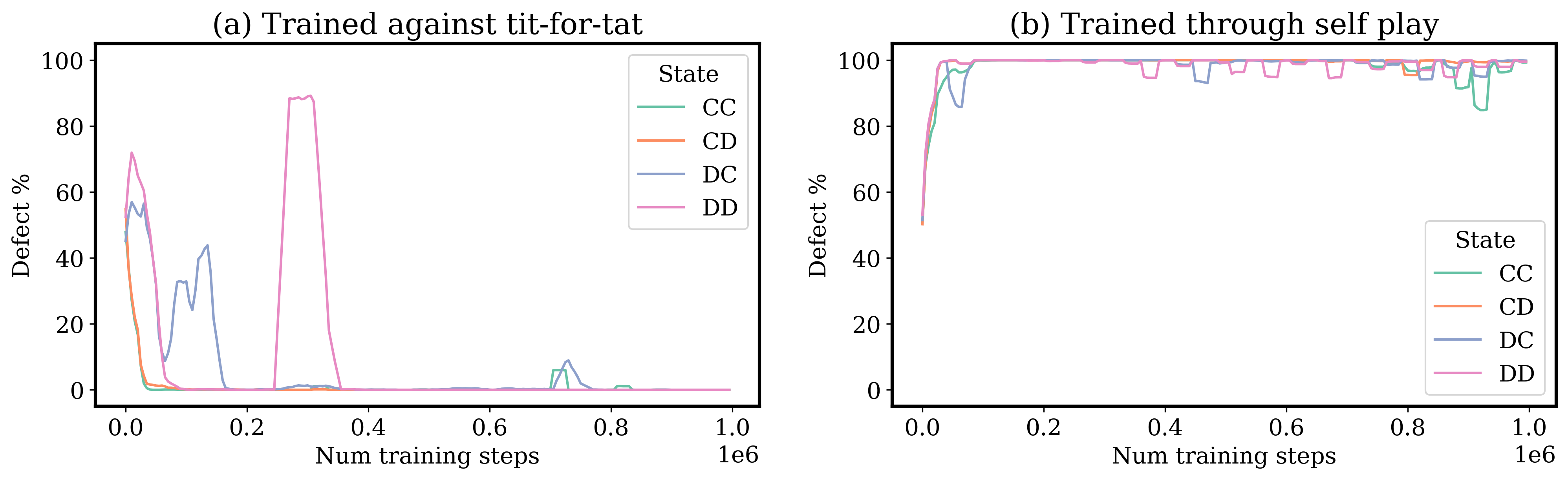}
  \caption{Policy (i.e. probability of playing defect for each state) throughout training against tit-for-tat and with self-play. In (a), as tit-for-tat will always reciprocate cooperation and never exploit it, the RL agent learns to cooperate regardless of the past outcome. While the resulting payoff for both agents is high, always cooperating is not a Nash and an exploitative agent could easily play defect to improve their payoff. In (b), despite the existence of higher payoff Nash equilibria, the RL agent (through self-play) converges quickly to a sub-optimal equilibrium of always defecting.}
  \label{fig:production_consumption_savings}
\end{figure}

\subsection{Cooperation with Fixed-Length Horizon Episodes}
While convenient, fixed horizons present potential theoretical/practical issues for the emergence of cooperation in prisoner's dilemma/free-rider type games. Specifically, for the case of the repeated prisoner's dilemma, cooperative equilibria exist \textit{but only for infinite/non-fixed horizons}~\citep{dawkins2016selfish,WinNT}. For a fixed horizon length $n$, the only possible equilibrium is one where agents defect at all timesteps. 

The gist of the issue is as follows. At the last timestep $n$, there is no future interaction between agents meaning the game reduces to the non-iterated prisoner's dilemma (where the equilibrium is known to be mutual defection). Any agent who chooses cooperate will simply be defected on by the other agent. However, given agents are perfectly rational, they all know that at the last step all agents will select defect. As such, their action at step $n-1$ will have no bearing on the future $\Rightarrow$ the game at step $n-1$ is once again a non-iterated prisoner's dilemma with mutual defection. Going backwards, by induction, we get that the subgame perfect equilibrium for the fixed-length iterated prisoner's dilemma must be defection at every step. Using horizon lengths drawn from some distribution (e.g., geometric distribution) could help mitigate against this issue (or at least against any weird behavior in the last period).

\subsection{Curriculum learning}
% Show that RL agents suboptimally solve a separate task that is not affected by the prisonner's dilemma problem
In addition to the general cooperation and negotiation challenge, agents also need to learn other elements of the world (the tradeoff between savings and consumption, the benefits of trade, the impact of mitigation on the environment, etc.). Learning these in conjunction with learning to cooperate/defect is challenging. As such, a curriculum learning scheme (similar to the one used in \cite{zheng2020ai}) could be beneficial (or for example having some set of pre-trained agents already familiar with the environment that get fine-tuned to specific negotiation protocols).

\section{Conclusion}
Ultimately, the use of MARL to approximate agent behavior in a complex economic model is a promising research direction. Due to the complexity of the model and the difficulties linked to cooperation for deep MARL, the learning aspect of the simulation is likely to be a significant challenge that leads to certain negotiation protocols being undervalued (agents not converging to high value equilibria) 
or overvalued (agents exploiting certain aspects of the simulation). It is also unclear to what extent it is important to converge to Nash equilibria: the outcome of human negotiation may not be optimal and it is possible that unconditional cooperation exists. Consequently, we suggest robustifying the evaluation criteria of solutions and further investigating the properties of MARL in such an environment to better understand the strengths and weaknesses of proposed negotiation protocols. 

\clearpage

\bibliography{refs}
\bibliographystyle{abbrvnat}

\clearpage
\appendix

% \section*{Appendix}

\section{Iterated Prisoner's Dilemma (IPD)}
\label{Appendix}
The setup of the iterated prisoner's dilemma is as follows: 2 agents must maximize their reward where at each step they play either \textbf{cooperate} (C) or \textbf{defect} (D). The rewards are setup so that both agents cooperating (CC) is better than both defecting (DD). However, given that it's partner is cooperating, it is better for that agent to betray its partner and defect (CD), in which case the partner will get a worse reward (DC) than if both defected.

\begin{table}[h]
\centering
    \label{tab:prisoner_dilemma}
    \centering
    \begin{tabular}{|c|c|c|}
      \toprule
      \textbf{Player 1 / Player 2} & \textbf{Cooperate} & \textbf{Defect} \\ 
      \midrule
      \textbf{Cooperate} & (4, 4) & (0, 5) \\  
      \textbf{Defect} & (5, 0) & (1, 1) \\
      \bottomrule
    \end{tabular}
    \vspace{1em}
    \caption{Example of a possible payoff matrix of a prisoner's dilemma type game. The first entry in each cell is the reward for Player 1 and the second is the reward for Player 2.}
\end{table}
\vspace{-1em}

While in the one step game, the only equilibria is both sides defecting, the case of the iterated game (i.e. where agents play multiple rounds in sequence) has the potential to have multiple more complex equilibria (depending on the value of the rewards ~\citep{WinNT}). Other than the always defect equilibrium, the others have the potential for cooperation and thus higher overall payoff.
\begin{itemize}
    \item \textbf{Always defect:} As in the one-step case, both agents always playing defect is an equilibria.
    \item \textbf{Tit-for-tat:} If the other agent played cooperate at the last step, play cooperate, otherwise defect.
    \item \textbf{Grim trigger:} Play cooperate until the other agent defects, then always defect.
    \item \textbf{Pavlov:} Play cooperate if you both cooperated at the last step. If your partner cooperates and you defected, keep trying to exploit them and defect. If you cooperated and your partner defected, defect as well (punish). Finally, if you both defected, try to revive cooperation by cooperating.
\end{itemize}

%%%%%%%%%%%%%%%%%%%%%%%%%%%%%%%%%%%%%%%%%%%%%%%%%%%%%%%%%%%%

\end{document}